\documentclass[10pt,twocolumn,letterpaper]{article}

\usepackage{cvpr}              

\usepackage{graphicx}
\usepackage{amsmath}
\usepackage{amssymb}
\usepackage{booktabs}
\usepackage{multirow}
\usepackage{subcaption}
\newcommand{\bhline}[1]{\noalign{\hrule height #1}}

\usepackage{algorithm}
\usepackage{algpseudocode}
\usepackage[accsupp]{axessibility}

\usepackage[pagebackref,breaklinks,colorlinks]{hyperref}

\usepackage[capitalize]{cleveref}
\crefname{section}{Sec.}{Secs.}
\Crefname{section}{Section}{Sections}
\Crefname{table}{Table}{Tables}
\crefname{table}{Tab.}{Tabs.}

\def\cvprPaperID{8} 
\def\confName{CVPR}
\def\confYear{2023}

\begin{document}

\title{One-shot and Partially-Supervised Cell Image Segmentation \\ Using Small Visual Prompt}

\author{Sota Kato\\
Meijo University\\
{\tt\small 150442030@ccalumni.meijo-u.ac.jp}
\and
Kazuhiro Hotta\\
Meijo University\\
{\tt\small kazuhotta@meijo-u.ac.jp}
}
\maketitle

\begin{abstract}
Semantic segmentation of microscopic cell images using deep learning is an important technique, however, it requires a large number of images and ground truth labels for training.
To address the above problem, we consider an efficient learning framework with as little data as possible, and we propose two types of learning strategies: One-shot segmentation which can learn with only one training sample, and Partially-supervised segmentation which assigns annotations to only a part of images.
Furthermore, we introduce novel segmentation methods using the small prompt images inspired by prompt learning in recent studies.
Our proposed methods use a pre-trained model based on only cell images and teach the information of the prompt pairs to the target image to be segmented by the attention mechanism, which allows for efficient learning while reducing the burden of annotation costs.
Through experiments conducted on three types of microscopic cell image datasets, we confirmed that the proposed method improved the Dice score coefficient (DSC) in comparison with the conventional methods.
Our code is available at \url{https://github.com/usagisukisuki/Oneshot-Part-CellSegmentation}.
\end{abstract}

\section{Introduction}
Semantic segmentation, which assigns a class label to each pixel in an image, is a crucial technique for image analysis in the fields of medicine \cite{schlemper2018cardiac,ebner2018automated,roy2018inherent,jha2020kvasir} and biology \cite{graham2018sams,joon2017nuclei}. 
It has become possible to obtain objective results automatically by using deep learning and various methods have been proposed \cite{shit2021cldice,cao2023swin,tragakis2023fully,hiramatsu2018cell,shibuya2022cell,fujii2021x}. 
However, it is necessary to require a large number of images and ground truth labels when we design a deep learning model.
Particularly, generating ground truth requires the knowledge of human experts and takes a lot of time and costs. 

In recent studies, to tackle the above problem, few-shot segmentation \cite{liu2020crnet,lang2022learning,chan2022res2,dawoud2023knowing} for less training data, zero-shot segmentation \cite{ding2022decoupling,bucher2019zero} for only inference without training, and semi-supervised segmentation \cite{kwon2022semi,wu2022cross}, which learns with a small number of supervised training samples, have been proposed. 
Additionally, the novel idea of prompt learning has been gaining popularity in the field of natural language processing \cite{brown2020language}. 
This idea is an inference method using large-scale pre-trained models, and it has been reported that it can achieve higher accuracy than conventional few-shot and zero-shot learning methods in the field of image recognition \cite{luddecke2022image,wang2022learning,zhou2022conditional,jia2022visual}. 

\begin{figure}[t]
\begin{center}
\includegraphics[scale=0.36]{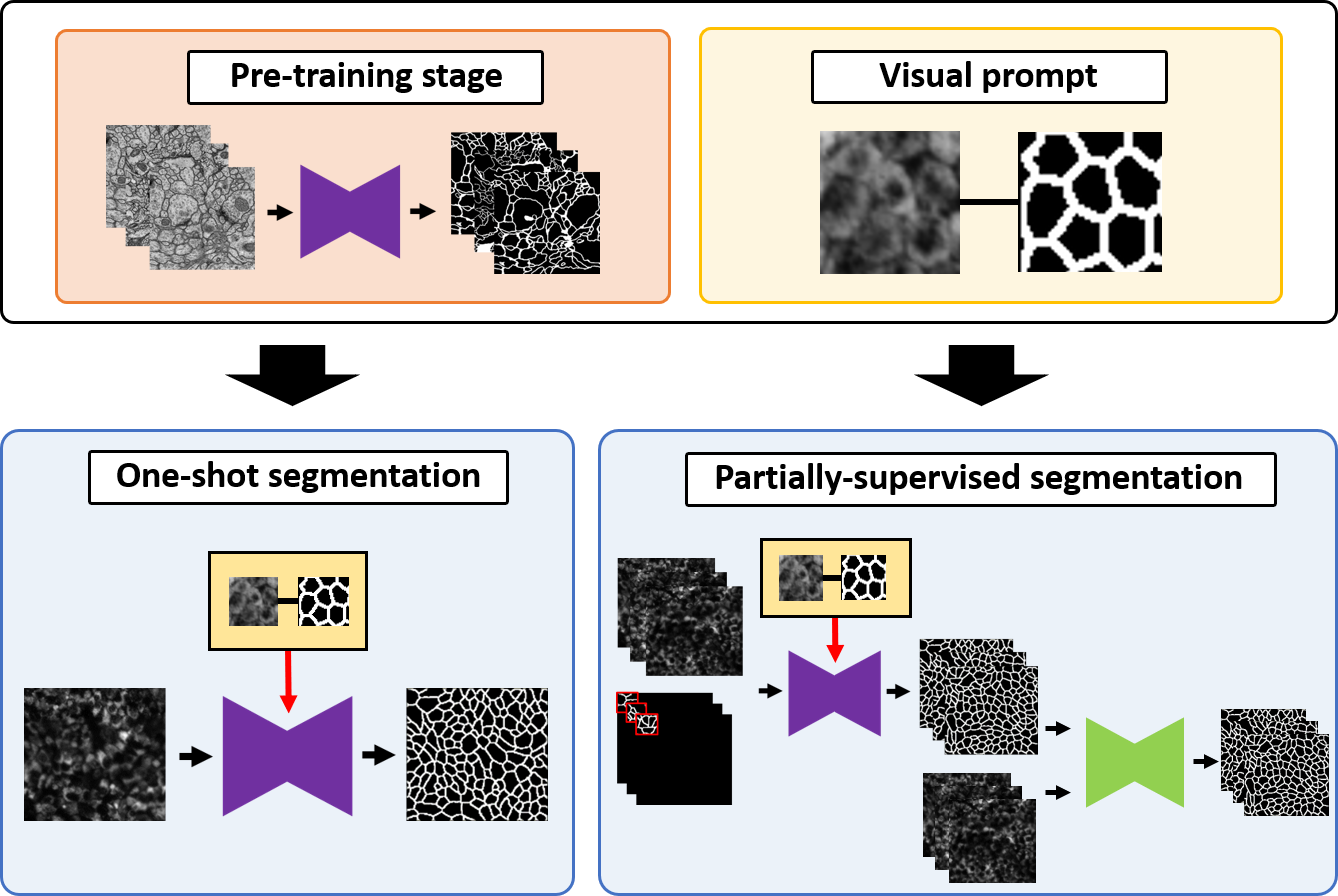}
\end{center}
\caption{Overview of our proposed strategies. 
Firstly, we build a pre-trained model using only cell images.
Secondly, we use the pre-trained model and the small visual prompt to learn one-shot segmentation and partially-supervised segmentation.}
\end{figure}

However, these approaches have been trained by natural images. 
As shown in Figure 2, it cannot be adapted well to images from other fields such as biology due to differences in domains between data.
Additionally, although conventional prompt learning methods often use both image and textual information \cite{liu2022open,wang2022learning,zhou2022conditional,luddecke2022image}, it is possible that general language models used by prompt learning cannot deal with specialized terms not included in the training data. 
In order to perform prompt learning to the data from different fields, we consider that it is necessary to use a pre-trained model, which is more specialized to the field.  

Therefore, we propose three strategies for cell image segmentation as shown in Figure 1.
Firstly, we build a pre-trained model using only cell image datasets.
By using the pre-trained model, we can learn more effectively even when training on another cell image dataset.
Secondly, we present novel strategies for one-shot segmentation and partially-supervised segmentation employing the above pre-trained model and visual prompts with small regions.
In one-shot segmentation, we assume learning with a single image and propose a novel learning method that utilizes a prompt image and label of a small region.
In partially-supervised segmentation, we
assume that a part of an image has annotation, and 
pseudo-labels are predicted using the same framework as one-shot segmentation, and segmentation network is trained with the pseudo labels.
Since cell images often have a fractal structure in which similar structures spread over the entire image, we consider that it is possible to segment the entire cell image by successfully using the information from small prompt regions.
Additionally, since labeling the prompt image with the small region is also easier than the entire image, it can reduce the annotation cost of human experts.

We evaluated our methods on three cell image datasets with different shapes.
From the experimental results, we confirmed the effectiveness of the proposed method in one-shot segmentation as well as the effectiveness of the pre-training model. 
Furthermore, the results by the proposed partially-supervised method and training on generated pseudo-labels demonstrated that the difference in accuracy between the proposed method and the method using the original dataset was within about 1.00\% for all cell image datasets. 
We have also confirmed that our proposed partially-supervised strategy can produce sufficient supervised labels with only part annotations.

This paper is organized as follows. 
Section 2 describes related works. 
Section 3 describes the details of the proposed methods. 
Section 4 shows the experimental results. 
Finally, we describe our summary and future works in Section 5.

The main contributions of this paper are as follows:
 \begin{itemize}
   \item We build a pre-trained model using only cell image datasets.
   By using this pre-trained model for one-shot segmentation and partially-supervised segmentation, we can create a model with more accuracy even with less data.
  \item We present a novel method for one-shot segmentation employing the pre-trained model and visual prompts with small regions.
  By using this method, we can achieve efficient learning even if only one training sample is used.
  \item Furthermore, we present a novel strategy for partially-supervised segmentation.
  We can generate pseudo labels from part labels using the same method as in one-shot segmentation, and we train the model again using these pseudo labels.
  By using this approach, we confirmed that the difference in accuracy between the proposed method and the method using all training samples in the original dataset was within about 1.00\% for all cell image datasets.  
 \end{itemize}

\begin{figure}[t]
    \centering
    \subfloat[General object image]{\includegraphics[clip, width=3.2in]{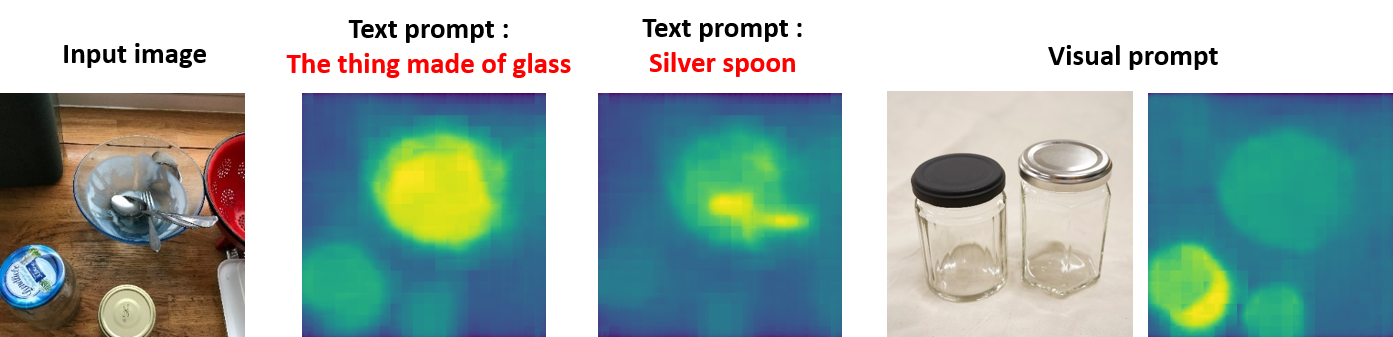}
        \label{overview of a training flow}}\\
    \vspace{0.2cm}
    \subfloat[ISBI2012 cell image \cite{globus_toolkit}]{\includegraphics[clip, width=3.2in]{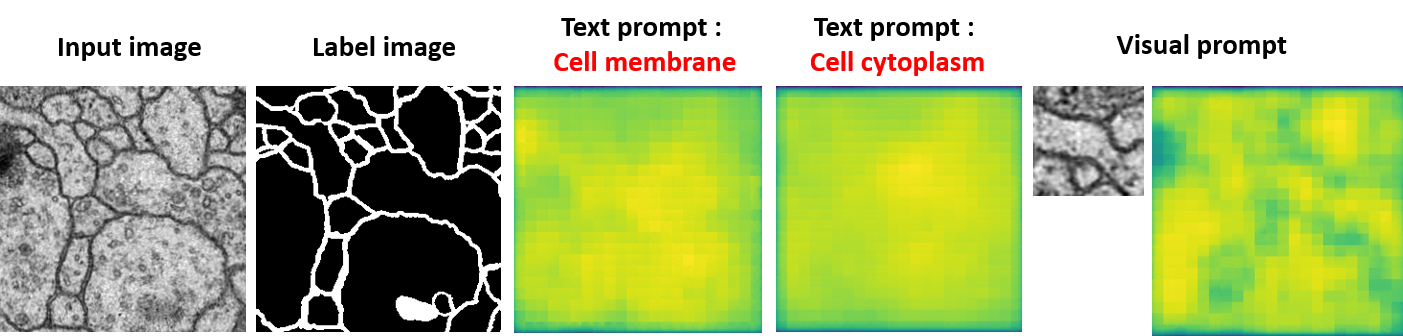}
        \label{overview of a inference flow}}\\
    \caption{Visualization results of CLIPSeg \cite{luddecke2022image}.}
\end{figure}

\section{Related works}
\subsection{One-shot segmentation}
Few-shot segmentation \cite{liu2020crnet,lang2022learning,li2021adaptive,chan2022res2,dawoud2023knowing,dawoud2021few} is a method that uses a small amount of training data. 
A query image of an unknown category, a support image of the same category, and a corresponding mask image are given as a support set for both training and inference. 
The goal is to learn efficiently with a small number of training data by using the information from the support set. 
In the field of biology, \cite{chan2022res2,dawoud2023knowing,dawoud2021few} have been proposed. 
Especially, when only one training image is used, it is called one-shot segmentation \cite{shaban2017one,raza2019weakly}, which is a more hard problem than few-shot segmentation.
Shaban et al. \cite{shaban2017one} propose a two-branched approach to one-shot semantic image segmentation taking inspiration from few-shot learning, and show significant improvements over the baselines on this benchmark.
Raza et al. \cite{raza2019weakly} present a simple yet effective approach, whereby exploiting information from base training classes in the conventional one-shot segmentation set-up allows for weak supervision to be easily used.

Furthermore, as an even more challenging problem, zero-shot segmentation has been proposed \cite{ding2022decoupling,bucher2019zero}. 
It is a method for segmenting unknown categories by using embedded features of words and similarity between features of the pre-trained model. 
Recently, the novel idea of prompt learning for zero-shot segmentation has been proposed \cite{luddecke2022image}.

Although numerous approaches have been proposed, there are a few one-shot segmentation methods for microscopic biology images.
We believe that one-shot segmentation is more necessary than few-shot segmentation because it reduces the burden of annotation cost furthermore.
Additionally, as shown in Figure 2, conventional zero-shot segmentation cannot work well due to the differences between data domains.
Therefore, we propose a novel method for one-shot cell image segmentation that can be improved segmentation accuracy in comparison with conventional approaches.

\subsection{Partially-supervised segmentation}

Semantic segmentation using deep learning requires a large number of annotations, and the performance decreases significantly when the number of annotations is small.
However, generating a large number of annotations is a hardship for human experts.
Therefore, semi-supervised learning, which maintains performance with a small number of annotations, has been attracting attention \cite{kwon2022semi,wang2022semi,wu2022cross,tseng2021dnetunet,zhou2020deep}. 
In the field of medical and biological imaging, \cite{tseng2021dnetunet,zhou2020deep} have been proposed, and furthermore, a similar problem setup, which is called partially-supervised segmentation, has also been proposed in a recent study.
In semi-supervised segmentation, the setting is to use a few pieces of data annotated throughout an image, whereas, in partially-supervised segmentation, only a part of the image is annotated.
Xu et al. \cite{xu2021partially} propose an active learning framework. 
This strategy is used to select the most uncertain partially-label image based on the probability predicted by the segmentation model until finishing selecting the maximum number of annotated images, and the MixUp augmentation is used to learn a proper visual representation of both the annotated and unannotated images.

We believe that annotating a part of the image is less burden than annotating the full image in the case of the cell image because it is very hard to annotate each individual cell.
Additionally, cell images often have a fractal structure and we consider partially-supervised segmentation is effective.
However, the conventional technique for partially-supervised segmentation \cite{xu2021partially} is difficult to learn because of too many active learning steps. 
Our proposed partially-supervised learning is simple, and it is possible to achieve an even level of accuracy with the original data with only two training stages.
Further, since our methods for one-shot and partially-supervised segmentations are nearly identical, our training strategy can be used for multiple tasks.

\section{Methodology}
\begin{figure*}[t]
\begin{center}
\includegraphics[scale=0.53]{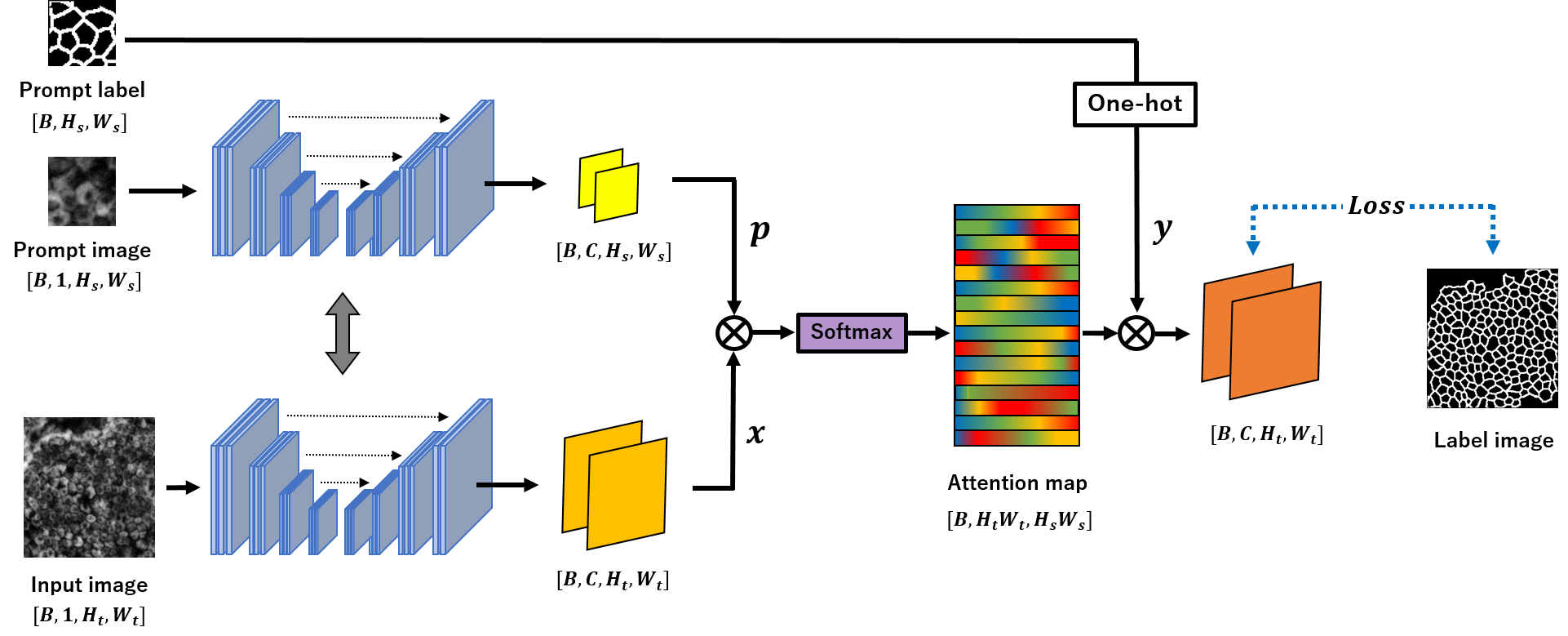}
\end{center}
\caption{Overview of the proposed method for one-shot segmentation.}
\end{figure*}
\begin{figure}[t]
\begin{center}
\includegraphics[scale=0.25]{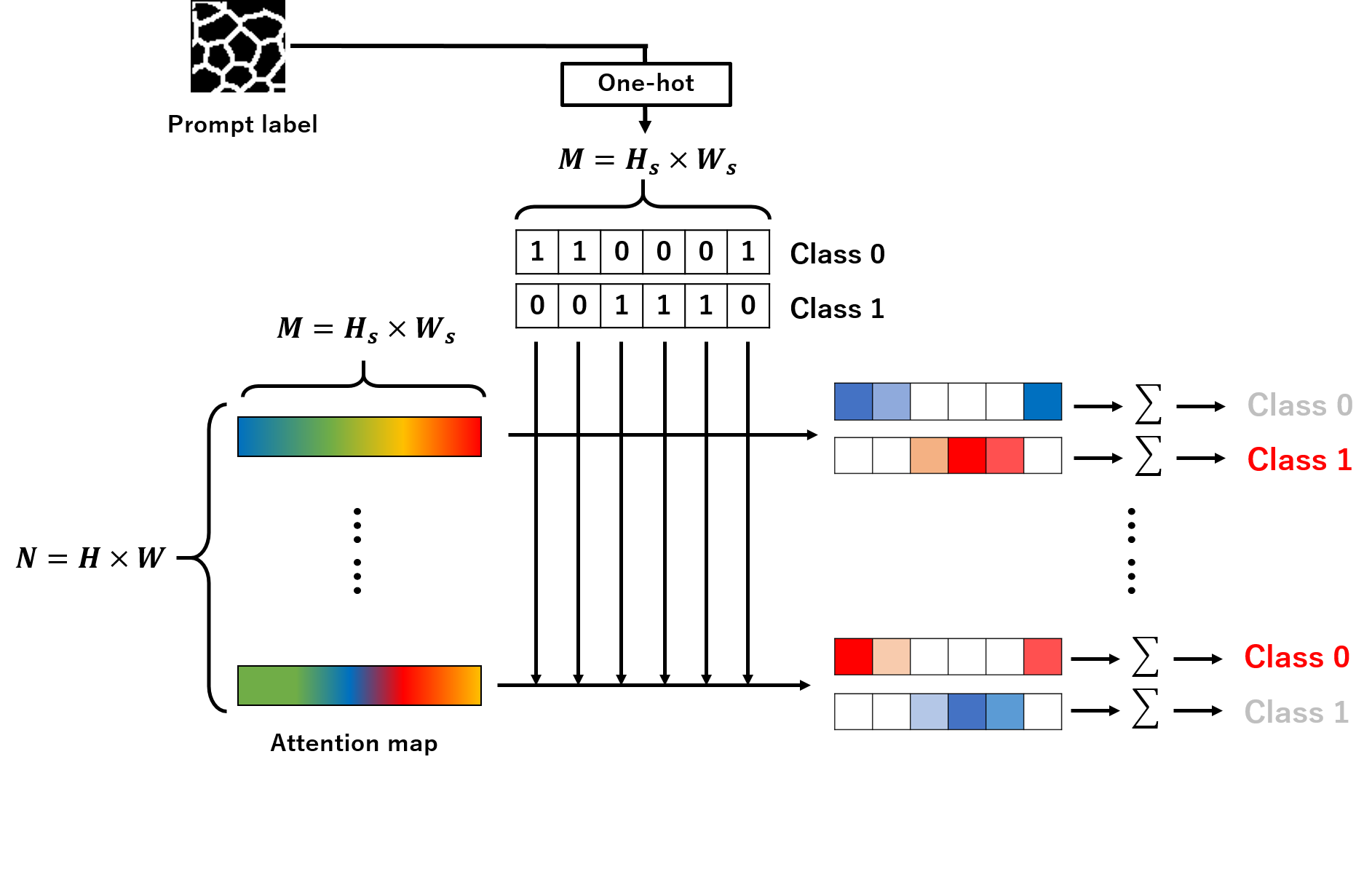}
\end{center}
\caption{Overview of methods for acquiring final output using the prompt label. Prompt label is converted to a one-hot label and the class with the highest value is output as a prediction from the inner product result with the attention map.}
\end{figure}
In Section 3, we present our novel approach for one-shot and partially-supervised segmentation for cell images.
By using these approaches, it is possible to learn with fewer annotations by simply preparing a visual prompt with a small region.

In Section 3.1, we present a detail of the pre-trained model. We present a novel architecture for one-shot learning in Section 3.2.
In Section 3.3, we present a novel learning strategy using pseudo labels under partially-supervised segmentation.

\subsection{Building the pre-trained model}

Firstly, we train a model using only cell image datasets. 
These datasets used in this study are ISBI2012 \cite{globus_toolkit}, ssTEM \cite{gerhard2013segmented}, and iRPE \cite{majurski2019cell} datasets. 
The network is U-Net \cite{ronneberger2015u} and the details of each dataset and training conditions are the same in Section 4.1. 
We build pre-trained models using two datasets other than the target dataset because of the fairness of the experiments.
For instance, for an evaluation experiment on ISBI2012, the pre-trained model using only ssTEM and iRPE datasets is adopted.

Consequently, three types of pre-trained models are built in this study. 
These models are used in our proposed strategies for one-shot segmentation in Section 3.2 and partially-supervised segmentation in Section 3.3.

\subsection{One-shot segmentation}
\begin{figure*}[t]
\begin{center}
\includegraphics[scale=0.36]{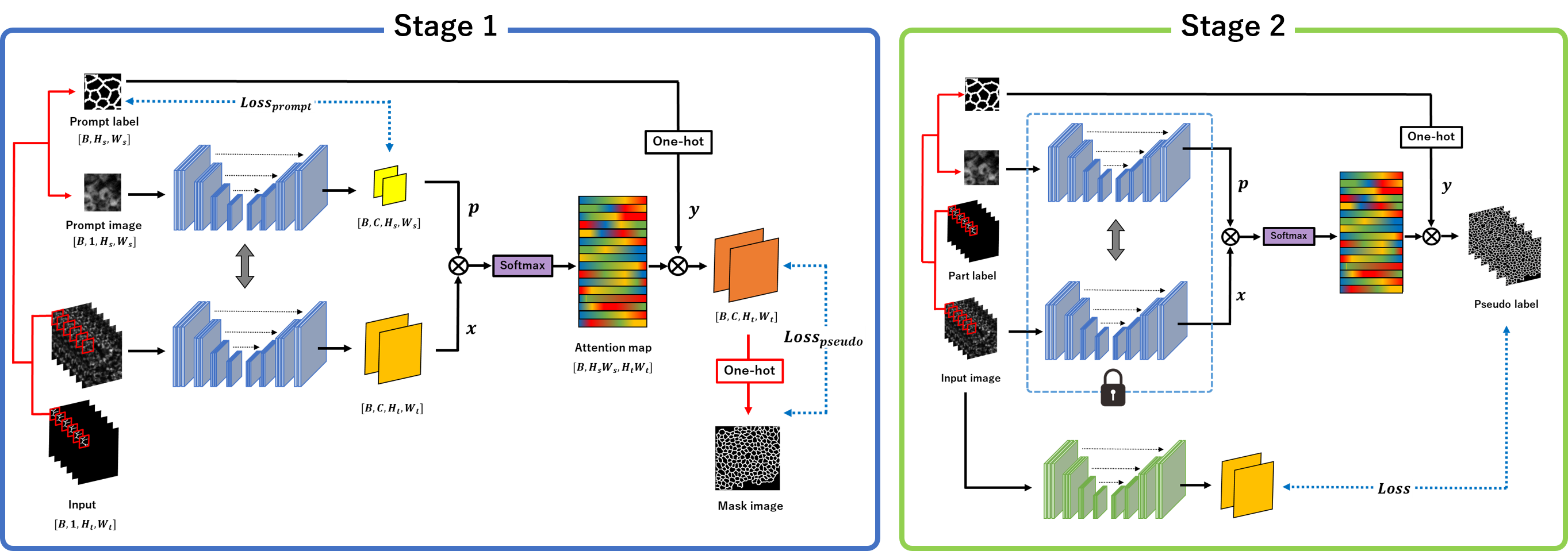}
\end{center}
\caption{Overview of our proposed strategies for partially-supervised segmentation.}
\end{figure*}

Figure 3 shows the overview of the proposed method for one-shot segmentation. 
In one-shot segmentation, we use three types of input data: the target microscopic cell image to be segmented, the prompt pairs of the small cell image, and the corresponding ground truth.
The target microscopic cell image and the prompt cell image are fed into the segmentation network.
The weights of two networks are shared, and the number of channels in output feature maps is the same as the number of classes in ground truth.
Final output features $\boldsymbol{x}\in\mathbb{R}^{C\times N}$ of the input image to be segmented and the final output $\boldsymbol{p}\in\mathbb{R}^{C\times M}$ of the prompt image are an inner product to generate a normalized attention map using a softmax function as
\begin{eqnarray}
  \beta_{i,j} = \frac{exp(s_{i,j})/ \tau}{\sum_j^M exp(s_{i,j})/ \tau}, \:\:\:where\:\: s_{i,j}=\boldsymbol{x_{c,i}}^\top \boldsymbol{p_{c,j}}
\end{eqnarray}
where $\beta_{i,j}$ is the attention map and represents the degree to which the $j$-th region is associated with the $i$-th region, $C$ is the number of classes, and $N=H_t \times W_t, M=H_s \times W_s$ are the number of feature locations from the final output feature maps.
$\tau$ is a temperature parameter that is to align probabilities for attention maps.
By using one-hot teacher label $\boldsymbol{q}\in\mathbb{R}^{C\times M}$ corresponding to the prompt cell image, the final output is $\boldsymbol{o=(o_1, o_2, ..., o_i, ..., o_N)}\in\mathbb{R}^{C\times N}$ and can be calculated in Equation (2). 
\begin{eqnarray}
  \boldsymbol{o_i} = \sum_j^M \beta_{i,j}\boldsymbol{q_j}^\top
\end{eqnarray}
where $\beta_{i,j}$ is the attention map calculated by the inner product of $\boldsymbol{x}$ and $\boldsymbol{p}$.
We can obtain the similarity between the output features from the input image and the prompt image in the attention map.
Subsequently, as shown in Figure 4, a final prediction can be performed by referring to the class labels of  output features of the prompt image that have high similarity to the output feature of the image to be segmented.

We employ the cross entropy loss in equation (3).
\begin{eqnarray}
    Loss = -\frac{1}{C}\sum_{c=1}^{C}\sum_{i=1}^{H_t\times W_t} y_i^c\log \sigma(o_i^c)
\end{eqnarray}
where $C$ is the number of classes, $y_i^c$ is the teacher labels associated with the input image, and $\sigma(o_i^c)$ is the probability value after a softmax function as $\sigma(o_i^c)=\frac{exp(o_i^c)}{\sum_{j} exp(o_i^j)}$. 
Further, $o_i^c$ is the $c$-th element of $\boldsymbol{o_i}$, which is a final output vector of the deep neural network.

Since the proposed method can utilize the information of the prompt image and their corresponding teacher labels for both training and inference, we consider that it can create a highly accurate model even with one-shot segmentation.

\subsection{Partially-supervised segmentation}

Figure 5 shows the overview of our proposed strategy for partially-supervised segmentation. 
In partially-supervised segmentation, it is assumed that only a part of the input cell image is given an annotation, and cannot be trained in the basic training method for segmentation using deep learning.
Therefore, the proposed strategy consists of two stages.
First, the network is trained to generate pseudo-labels using only part of the label information.
Second, another segmentation network is trained using generated pseudo-labels in a full-scratch learning.

In the stage of learning to generate pseudo-labels, we crop the part region of a cell image which is assigned annotation, and feed it with the partial label into the network as the prompt pairs. 
Consequently, we use three types of input to the segmentation network in the same way as in Section 3.2. 
As in the proposed one-shot segmentation, the final output is obtained by referring to the ground truth corresponding to the prompt image using the attention architecture. 
Then, the final output is translated a mask label with one-hot type by the argmax function, and the network trains using the mask label as a self-supervised learning. 
Additionally, it trains bringing the output from the prompt image closer to the prompt label.

We design loss functions in Equation (4)-(6).
We use the cross-entropy loss between the pseudo mask labels and the final predictions in Equation (5), and the cross-entropy loss between the prediction regions for the prompt image and the prompt label in Equation (6). 
As the final loss function, we employ in combination in Equation (4).
\begin{eqnarray}
    Loss &=& Loss_{pseudo} + Loss_{prompt}\\
    Loss_{pseudo} &=& -\frac{1}{C}\sum_{c=1}^{C}\sum_{i=1}^{H_t\times W_t} m_i^c\log \sigma(o_i^c)\\
    Loss_{prompt} &=& -\frac{1}{C}\sum_{c=1}^{C}\sum_{i=1}^{H_s\times W_s} q_i^c\log \sigma(p_i^c)
\end{eqnarray}
where $C$ is the number of classes, $m_i$ is the pseudo mask label, $q_i$ is the prompt label, $\sigma(p_i^c)$ is the probability value from the input image, and $\sigma(p_i^c)$ is the probability value from the prompt image after a softmax function as $\sigma(p_i^c)=\frac{exp(p_i^c)}{\sum_{j} exp(p_i^j)}$, further, $p_i^c$ is the $c$-th element of $\boldsymbol{p_i}$, which is a final output vector of the deep neural network.
In one-shot segmentation, we do not use the loss for the prompt image since we use only one prompt pair and there is a possibility of over-fitting.
However, in partially-supervised segmentation, we add $Loss_{prompt}$ to improve the quality of the pseudo label because we can use different prompt pairs during learning.

In the stage of training using pseudo-labels, we crop the part region of the cell image, which is assigned annotation, and feed it with the partial label into the network as the prompt pairs. 

By using the proposed strategy, even when only some of the teachers are given, the proposed method can extend the information of the sample images to the whole image by using the attention structure, which enables segmentation of the whole image.

\section{Experiments}
\begin{table*}[t]
    \centering
    \caption{Comparison results for one-shot segmentation.}
    \scalebox{0.78}{
    \begin{tabular*}{22cm}{@{\extracolsep{\fill}}rccccccccc} \bhline{1.0pt}
    \multicolumn{1}{r}{} & \multicolumn{3}{c}{ISBI2012 \cite{globus_toolkit}} & \multicolumn{3}{c}{ssTEM \cite{gerhard2013segmented}}& \multicolumn{3}{c}{iRPE \cite{majurski2019cell}}\\
    \cmidrule(lr){2-4}%
    \cmidrule(lr){5-7}%
    \cmidrule(lr){8-10}%
    \multicolumn{1}{r}{DSC}&Average&Background&Membrane &Average&Background&Membrane&Average&Background&Membrane\\
    \hline
        \multicolumn{1}{r}{$Original\:\:learning$}&&&&&&&&&\\
    \hline
        U-Net (Full-scratch) &86.66\tiny{±0.50}&94.04\tiny{±0.15}&79.28\tiny{±0.86}&91.60\tiny{±0.14}&96.54\tiny{±0.05}&86.65\tiny{±0.23}&74.39\tiny{±0.19}&84.60\tiny{±0.34}&64.17\tiny{±0.67}\\
        U-Net (Pre-trained)&86.92\tiny{±0.40}&94.15\tiny{±0.20}&79.70\tiny{±0.61}&90.99\tiny{±0.19}&96.28\tiny{±0.12}&85.70\tiny{±0.27}&73.70\tiny{±0.11}&84.00\tiny{±0.21}&63.41\tiny{±0.38}\\
        \hline
        \multicolumn{1}{r}{$One\mathchar`-shot\:\:learning$}&&&&&&&&&\\
        \hline
        U-Net (Full-scratch)&78.80\tiny{±1.73}&89.83\tiny{±0.86}&67.77\tiny{±2.67}&78.74\tiny{±2.46}&92.54\tiny{±0.26}&64.94\tiny{±4.66}&59.25\tiny{±0.05}&78.96\tiny{±0.44}&39.53\tiny{±0.40}\\
        U-Net (Pre-trained)&81.01\tiny{±1.86}&91.51\tiny{±1.06}&70.51\tiny{±2.69}&82.42\tiny{±0.71}&92.71\tiny{±0.36}&72.13\tiny{±1.07}&53.85\tiny{±1.44}&79.77\tiny{±0.61}&27.93\tiny{±2.28}\\
        Co-FCN \cite{raza2019weakly}&78.19\tiny{±2.34}&89.87\tiny{±1.62}&66.51\tiny{±3.48}&83.61\tiny{±0.24}&93.14\tiny{±0.07}&74.09\tiny{±0.46}&54.81\tiny{±3.57}&\textbf{81.88\tiny{±0.32}}&27.75\tiny{±7.38}\\
        OSLSM \cite{shaban2017one}&78.67\tiny{±1.43}&90.03\tiny{±0.58}&67.32\tiny{±2.29}&84.00\tiny{±0.83}&93.09\tiny{±0.50}&74.91\tiny{±1.17}&62.94\tiny{±0.76}&79.52\tiny{±0.72}&46.36\tiny{±2.25}\\
        FSMICS (one-shot) \cite{dawoud2021few}&79.85\tiny{±0.93}&90.67\tiny{±0.43}&69.03\tiny{±1.42}&83.42\tiny{±0.10}&92.82\tiny{±0.17}&74.02\tiny{±0.30}&61.14\tiny{±2.18}&79.48\tiny{±0.30}&42.79\tiny{±4.65}\\
        \hline
        Ours (Full-scratch)&80.18\tiny{±1.26}&91.11\tiny{±0.66}&69.25\tiny{±1.87}&83.13\tiny{±0.26}&92.30\tiny{±0.25}&73.95\tiny{±0.29}&\textbf{64.00\tiny{±1.33}}&78.85\tiny{±0.72}&\textbf{49.16\tiny{±2.03}}\\
        Ours (Pre-trained)&\textbf{81.25\tiny{±1.55}}&\textbf{91.68\tiny{±1.04}}&\textbf{70.81\tiny{±2.18}}&\textbf{85.47\tiny{±0.08}}&\textbf{94.41\tiny{±0.01}}&\textbf{76.53\tiny{±0.15}}&63.85\tiny{±1.53}&78.56\tiny{±1.69}&49.15\tiny{±1.61}\\
    \bhline{1.0pt} 
    \end{tabular*}
    }
\end{table*}
\begin{table*}[t]
    \centering
    \caption{Comparison results for partially-supervised segmentation.}
    \scalebox{0.78}{
    \begin{tabular*}{22cm}{@{\extracolsep{\fill}}rccccccccc} \bhline{1.0pt}
    \multicolumn{1}{r}{} & \multicolumn{3}{c}{ISBI2012 \cite{globus_toolkit}} & \multicolumn{3}{c}{ssTEM \cite{gerhard2013segmented}}& \multicolumn{3}{c}{iRPE \cite{majurski2019cell}}\\
    \cmidrule(lr){2-4}%
    \cmidrule(lr){5-7}%
    \cmidrule(lr){8-10}%
    \multicolumn{1}{r}{DSC} & Average & Background & Membrane & Average & Background & Membrane  & Average & Background & Membrane \\
    \hline
        \multicolumn{1}{r}{$Original\:\:learning$}&&&&&&&&&\\
    \hline
        U-Net (Full-scratch) &86.66\tiny{±0.50}&94.04\tiny{±0.15}&79.28\tiny{±0.86}&91.60\tiny{±0.14}&96.54\tiny{±0.05}&86.65\tiny{±0.23}&74.39\tiny{±0.19}&84.60\tiny{±0.34}&64.17\tiny{±0.67}\\
        U-Net (Pre-trained)&86.92\tiny{±0.40}&94.15\tiny{±0.20}&79.70\tiny{±0.61}&90.99\tiny{±0.19}&96.28\tiny{±0.12}&85.70\tiny{±0.27}&73.70\tiny{±0.11}&84.00\tiny{±0.21}&63.41\tiny{±0.38}\\
        \hline
        \multicolumn{1}{r}{$Pseudo\:\:label\:\:learning$}&&&&&&&&&\\
        \hline
        Ours (Full-scratch) &85.94\tiny{±0.28}&94.03\tiny{±0.13}&77.85\tiny{±0.43}&89.78\tiny{±0.02}&95.80\tiny{±0.05}&83.76\tiny{±0.05}&\textbf{74.35\tiny{±0.12}}&\textbf{85.21\tiny{±0.10}}&\textbf{63.49\tiny{±0.25}}\\
        Ours (Pre-trained)&\textbf{86.27\tiny{±0.32}}&\textbf{94.10\tiny{±0.18}}&\textbf{78.43\tiny{±0.46}}&\textbf{90.53\tiny{±0.12}}&\textbf{96.12\tiny{±0.06}}&\textbf{84.94\tiny{±0.18}}&73.30\tiny{±0.25}&84.18\tiny{±0.20}&62.41\tiny{±0.71}\\
    \bhline{1.0pt} 
    \end{tabular*}
    }
\end{table*}

\subsection{Datasets}

We used 2D electron microscopy images of the ISBI2012 challenge (ISBI2012) \cite{globus_toolkit}, serial sectioning transmission electron microscopy (ssTEM) \cite{gerhard2013segmented}, and absorbance microscopy images of human iRPE cells (iRPE) \cite{majurski2019cell} as datasets. 
All datasets are for binary segmentation of tubular structures spread over an image, i.e., cell membrane and background.
The number of iRPE images is $1,032$ and the pixel count is $256 \times 256$.  
Since the resolution of ssTEM image is $1,024\times 1,024$ and the resolution of ISBI2012 image is $512\times 512$, and we cropped a region of $256\times 256$ pixels from the original images due to the limitation of GPU memory.
There is no overlap for cropping areas, and consequently, the total number of crops is 320 in ssTEM and 120 in ISBI2012.
Figure 6 shows the examples in the datasets.

We randomly rearranged the images. 
Afterward, we divided each dataset into 2 to 1 in index order and prepared them as training or inference data, and used three-fold cross validation while switching the training and inference data.

In the experiment of one-shot segmentation, we used only one fixed training image from the training data, and in inference, we used the original inference data.
The prompt image was selected randomly from the training data with a different image other than the training image, and the fixed coordinates were cropped.
Consequently, different prompt pairs were used for each 3-fold cross validation.

In the experiment of partially-supervised segmentation, all labels in the training data were masked by filling in zeros except for the fixed coordinates, and we used the region where the label exists as prompt images.
The size of the prompt was set to $64\times 64$ pixels in all experiments.
Ablations of different prompt sizes are shown in Section 4.3.

\begin{figure}[t]
    \begin{tabular}{ccc}
      \begin{minipage}{0.31\hsize}
        \centering
        \includegraphics[scale=0.31]{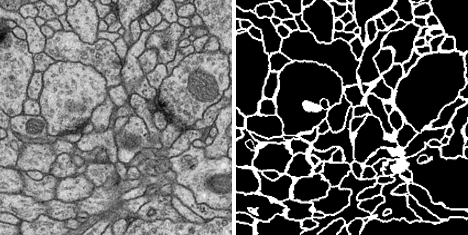}
        \subcaption{}
    \end{minipage}%
    \begin{minipage}{0.31\hsize}
        \centering
        \includegraphics[scale=0.31]{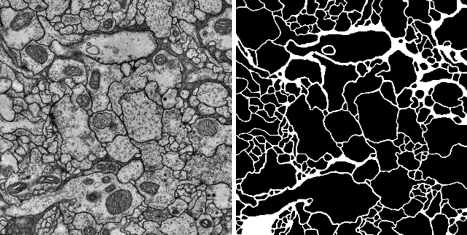}
        \subcaption{}
    \end{minipage}%
    \begin{minipage}{0.31\hsize}
        \centering
        \includegraphics[scale=0.31]{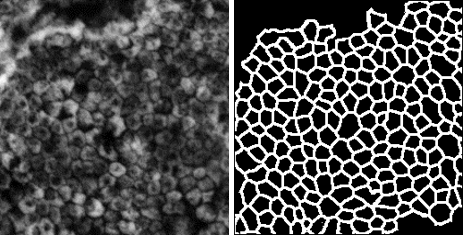}
        \subcaption{}
    \end{minipage}%
    \end{tabular}
    \caption{Examples of datasets. (a) ISBI2012 dataset \cite{globus_toolkit}, (b) ssTEM dataset \cite{gerhard2013segmented}, and (c) iRPE dataset \cite{majurski2019cell}. All datasets are labeled as: cell membrane (white) and background (black).}
\end{figure}

\subsection{Training conditions}

The batch size for training was set to 4, and we used Adam ($betas = 0.9, 0.999$) for optimization. 
The initial learning rate was 0.001, and we used a learning rate schedule that decays the learning rate by $0.1$ at 180 epoch and again at 190 epoch.
For data pre-processing, training samples were flipped horizontally, rotated with an angle randomly selected within $\theta$ = $-90^\circ$ to $90^\circ$, and normalized from zero to one.
For inference, images were normalized from zero to one.
All experiments were conducted using a three-fold cross validation and we applied the Dice score coefficient (DSC) as evaluation metric.
The average DSC from three validations was used for evaluation. 
We used a single Nvidia RTX Quadro 8000 GPU as a calculator.


\subsection{Experimental results}
\begin{figure*}[t]
    \centering
    \begin{tabular}{ccccc}
      \begin{minipage}{0.19\hsize}
        \includegraphics[scale=0.5]{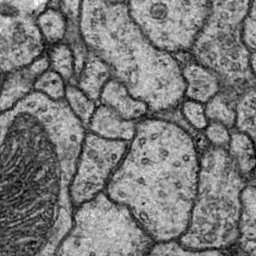}
    \end{minipage}%
    \begin{minipage}{0.19\hsize}
        \includegraphics[scale=0.5]{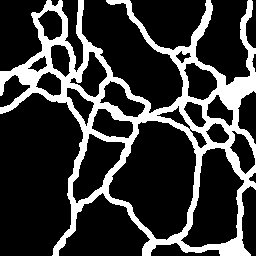}
    \end{minipage}%
    \begin{minipage}{0.19\hsize}
        \includegraphics[scale=0.5]{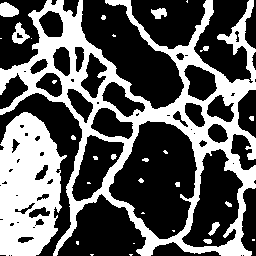}
    \end{minipage}%
    \begin{minipage}{0.19\hsize}
        \includegraphics[scale=0.5]{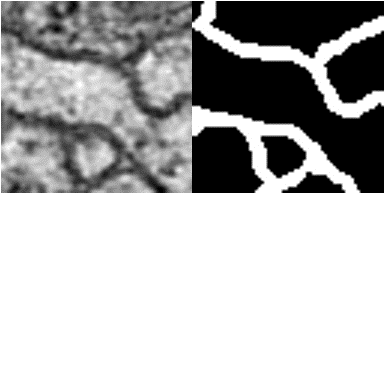}
    \end{minipage}%
    \begin{minipage}{0.19\hsize}
        \includegraphics[scale=0.5]{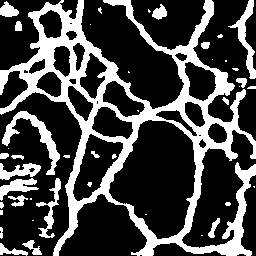}
    \end{minipage}%
    \end{tabular}
    
    \vspace{0.4cm}
    
    \begin{tabular}{ccccc}
      \begin{minipage}{0.19\hsize}
        \includegraphics[scale=0.5]{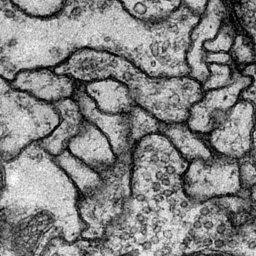}
    \end{minipage}%
    \begin{minipage}{0.19\hsize}
        \includegraphics[scale=0.5]{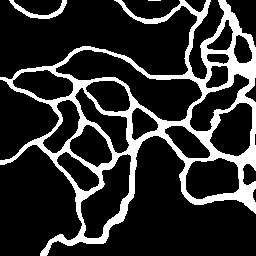}
    \end{minipage}%
    \begin{minipage}{0.19\hsize}
        \includegraphics[scale=0.5]{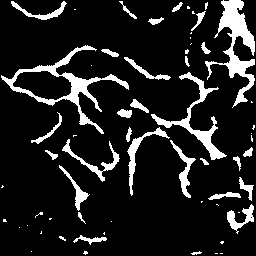}
    \end{minipage}%
    \begin{minipage}{0.19\hsize}
        \includegraphics[scale=0.5]{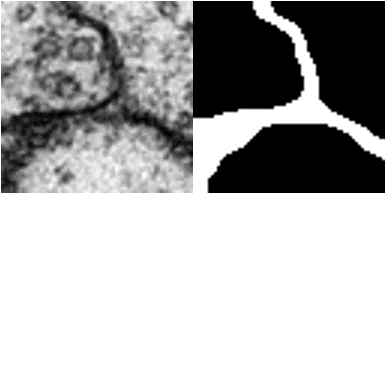}
    \end{minipage}%
    \begin{minipage}{0.19\hsize}
        \includegraphics[scale=0.5]{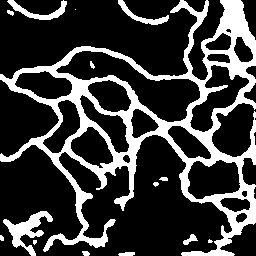}
    \end{minipage}%
    \end{tabular}
    
    \vspace{0.4cm}
    
    \begin{tabular}{ccccc}
      \begin{minipage}{0.19\hsize}
        \includegraphics[scale=0.5]{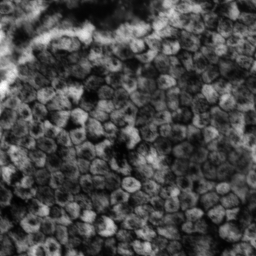}
        \subcaption{}
    \end{minipage}%
    \begin{minipage}{0.19\hsize}
        \includegraphics[scale=0.5]{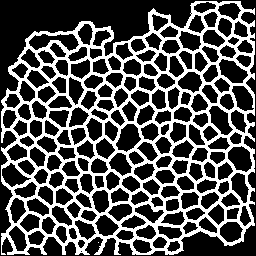}
        \subcaption{}
    \end{minipage}%
    \begin{minipage}{0.19\hsize}
        \includegraphics[scale=0.5]{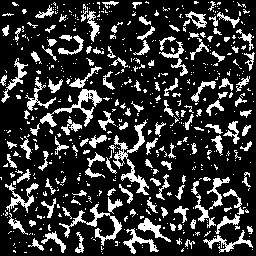}
        \subcaption{}
    \end{minipage}%
    \begin{minipage}{0.19\hsize}
        \includegraphics[scale=0.5]{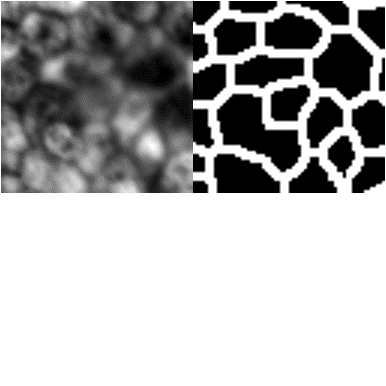}
        \subcaption{}
    \end{minipage}%
    \begin{minipage}{0.19\hsize}
        \includegraphics[scale=0.5]{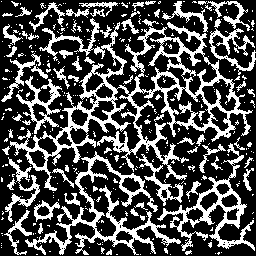}
        \subcaption{}
    \end{minipage}%
    \end{tabular}
    \caption{Qualitative results on one-shot segmentation. The top row is for the ISBI2012, the middle row is for the ssTEM, and the bottom row is for the iRPE dataset. (a) Input image, (b) Label image, (c) Original U-Net, (d) Prompt images ($64 \times 64$ pixels), and (e) Ours.}
\end{figure*}

\subsubsection{One-shot segmentation}
Table 1 shows comparison results for one-shot segmentation on the three datasets. 
As the comparison method, we employed conventional methods for one-shot segmentation \cite{raza2019weakly,shaban2017one,dawoud2021few}.
Original learning in Table 1 indicates the case when the model is trained by full training images, and one-shot learning indicates the case where it is only one training image.
Full-scratch in Table 1 indicates the case where the pre-trained model is not used, and pre-trained indicates the case where it is used.
The bold letters show the best DSC.
By using our proposed method, the highest average DSC was achieved for all datasets. 
Particularly, when we used the pre-trained model, the average DSC was improved by over 2.45\% for ISBI2012, 6.73\% for ssTEM, and 4.75\% for iRPE in comparison with a baseline using the original U-Net.
Furthermore, our method was more improved DSC than conventional approaches for one-hot segmentation.
When the iRPE was used, the model with full-scratch was more accurate than the model with pre-trained.
We consider that this may be because the iRPE cell image had a different structure compared to the two other cell images.

Figure 7 shows qualitative comparisons of one-shot segmentation through visualizations of three types of cell image datasets when we used a full-scratch learning.
As shown in Figure 7, the proposed method could segment the cell membrane that U-Net could not segment well. 
These results demonstrated the effectiveness of our method using the small visual prompt.

Table 3 shows an ablation study of one-shot segmentation when we evaluated various $\tau$ parameters, which are used in the softmax function of the attention mechanism, and smaller prompt sizes.
We evaluated $\tau=0.01,0.1,1.0,2.0$ as a temperature parameter for attention maps and further evaluated $32\times 32$ pixels of the prompt image, which is half size of $64\times 64$.
Comparison results with various temperature parameters demonstrated that adequate parameters depended on the data, and we consider that this is influenced by the thickness and complexity of the cell membrane to be segmented.
Our proposed method can be adapted to a variety of cell images by setting appropriate $\tau$. 
Furthermore, the accuracy did not decrease much with smaller prompt sizes.
By using smaller prompt images, we can further reduce the burden on the annotating cost.

\subsubsection{Partially-supervised segmentation}
\begin{figure}[t]
    \centering
    \begin{tabular}{ccc}
      \begin{minipage}{0.31\hsize}
        \includegraphics[scale=0.38]{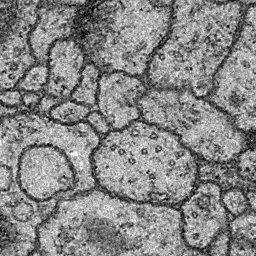}
    \end{minipage}%
    \begin{minipage}{0.31\hsize}
        \includegraphics[scale=0.38]{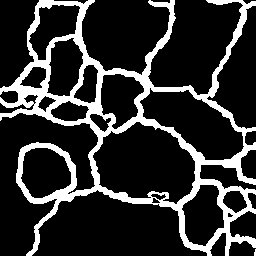}
    \end{minipage}%
    \begin{minipage}{0.31\hsize}
        \includegraphics[scale=0.38]{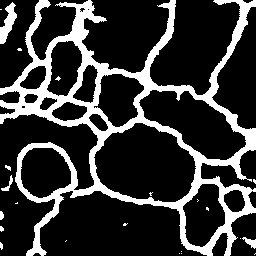}
    \end{minipage}%
    \end{tabular}
    
    \vspace{0.3cm}
        
    \begin{tabular}{ccc}
      \begin{minipage}{0.31\hsize}
        \includegraphics[scale=0.38]{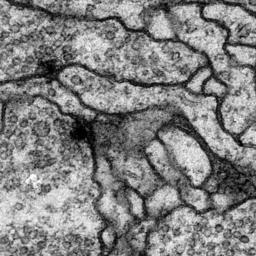}
    \end{minipage}%
    \begin{minipage}{0.31\hsize}
        \includegraphics[scale=0.38]{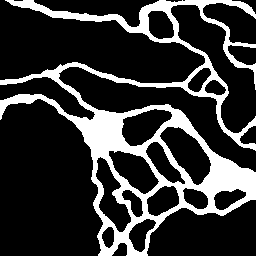}
    \end{minipage}%
    \begin{minipage}{0.31\hsize}
        \includegraphics[scale=0.38]{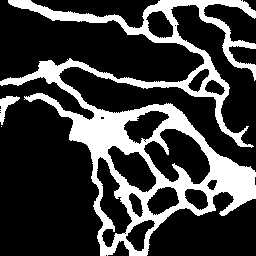}
    \end{minipage}%
    \end{tabular}
    
    \vspace{0.3cm}

    \begin{tabular}{ccc}
      \begin{minipage}{0.31\hsize}
        \includegraphics[scale=0.38]{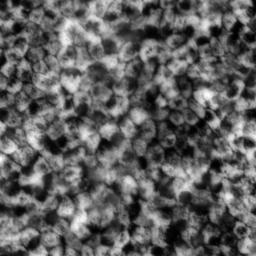}
        \subcaption{}
    \end{minipage}%
    \begin{minipage}{0.31\hsize}
        \includegraphics[scale=0.38]{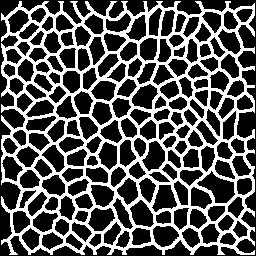}
        \subcaption{}
    \end{minipage}%
    \begin{minipage}{0.31\hsize}
        \includegraphics[scale=0.38]{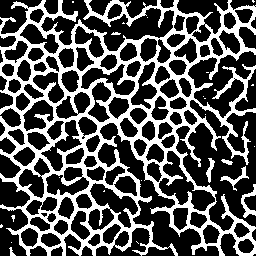}
        \subcaption{}
    \end{minipage}%
    \end{tabular}
    \caption{Segmentation results on partially-supervised segmentation. The top row is for the ISBI2012, the middle row is for the ssTEM, and the bottom row is for the iRPE dataset. (a) Input image, (b) Label image, and (c) Ours.}
\end{figure}
\begin{figure}[t]
    \centering
    \begin{tabular}{cccc}
    \begin{minipage}{0.23\hsize}
        \includegraphics[scale=0.28]{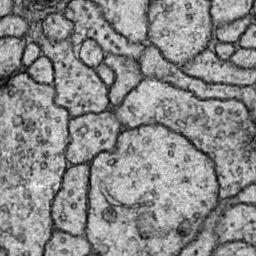}
    \end{minipage}%
    \begin{minipage}{0.23\hsize}
        \includegraphics[scale=0.28]{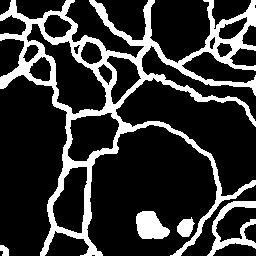}
    \end{minipage}%
    \begin{minipage}{0.23\hsize}
        \includegraphics[scale=0.28]{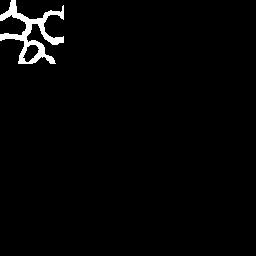}
    \end{minipage}%
    \begin{minipage}{0.23\hsize}
        \includegraphics[scale=0.28]{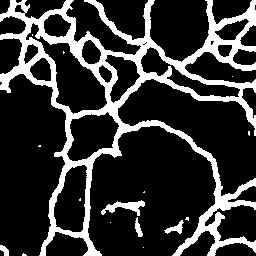}
    \end{minipage}%
    \end{tabular}
    
    \vspace{0.3cm}
        
    \begin{tabular}{cccc}
    \begin{minipage}{0.23\hsize}
        \includegraphics[scale=0.28]{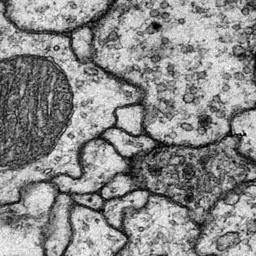}
    \end{minipage}%
    \begin{minipage}{0.23\hsize}
        \includegraphics[scale=0.28]{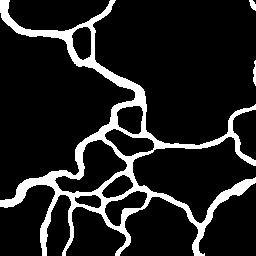}
    \end{minipage}%
    \begin{minipage}{0.23\hsize}
        \includegraphics[scale=0.28]{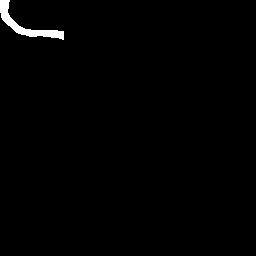}
    \end{minipage}%
    \begin{minipage}{0.23\hsize}
        \includegraphics[scale=0.28]{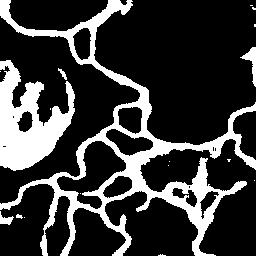}
    \end{minipage}%
    \end{tabular}
    
    \vspace{0.3cm}

    \begin{tabular}{cccc}
    \begin{minipage}{0.23\hsize}
        \includegraphics[scale=0.28]{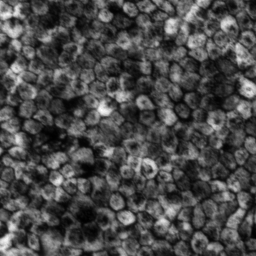}
        \subcaption{}
    \end{minipage}%
    \begin{minipage}{0.23\hsize}
        \includegraphics[scale=0.28]{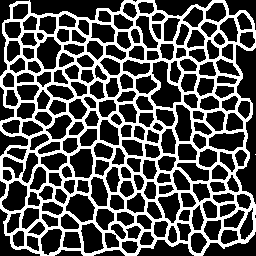}
        \subcaption{}
    \end{minipage}%
    \begin{minipage}{0.23\hsize}
        \includegraphics[scale=0.28]{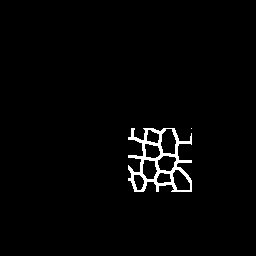}
        \subcaption{}
    \end{minipage}%
    \begin{minipage}{0.23\hsize}
        \includegraphics[scale=0.28]{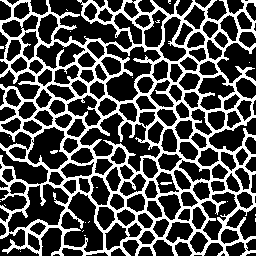}
        \subcaption{}
    \end{minipage}%
    \end{tabular}
    \caption{Qualitative results on partially-supervised segmentation. The top row is for the ISBI2012, the middle row is for the ssTEM, and the bottom row is for the iRPE dataset. (a) Training image, (b) Original label image, (c) Partial label image ($64\times 64$ pixels), and d) Generated pseudo label image in Stage 1 of Figure 5.}
\end{figure}
\begin{table}[t]
    \centering
    \caption{Ablation study of one-shot segmentation.}
    \scalebox{0.68}{
    \begin{tabular*}{12cm}{@{\extracolsep{\fill}}rcccc} \bhline{1.0pt}
    \multicolumn{1}{r}{Methods} & \multicolumn{1}{r}{Prompt size} & \multicolumn{1}{c}{ISBI2012 \cite{globus_toolkit}} & \multicolumn{1}{c}{ssTEM \cite{gerhard2013segmented}}& \multicolumn{1}{c}{iRPE \cite{majurski2019cell}}\\
    \hline
        $\tau$=0.01 (Full-scratch) &\multirow{4}{*}{$64\times 64$}&69.39\tiny{±12.16}&81.81\tiny{±0.82}&59.01\tiny{±7.15}\\
        $\tau$=0.1 (Full-scratch) &&78.61\tiny{±1.80}&83.12\tiny{±0.39}&\textbf{64.00\tiny{±1.33}}\\
        $\tau$=1.0 (Full-scratch) &&79.94\tiny{±1.21}&83.13\tiny{±0.26}&62.73\tiny{±1.31}\\
        $\tau$=2.0 (Full-scratch) &&80.18\tiny{±1.26}&83.12\tiny{±0.41}&63.31\tiny{±1.21}\\
        \hline
        $\tau$=0.01 (Pre-trained) &\multirow{4}{*}{$64\times 64$}&76.86\tiny{±4.24}&\textbf{85.47\tiny{±0.08}}&58.88\tiny{±6.61}\\
        $\tau$=0.1 (Pre-trained) &&81.13\tiny{±2.05}&85.40\tiny{±0.09}&58.13\tiny{±5.25}\\
        $\tau$=1.0 (Pre-trained) &&\textbf{81.25\tiny{±1.55}}&81.73\tiny{±1.09}&62.24\tiny{±1.09}\\
        $\tau$=2.0 (Pre-trained) &&81.15\tiny{±1.52}&84.18\tiny{±0.37}&63.85\tiny{±1.53}\\
    \hline
        $\tau$=0.01 (Full-scratch) &\multirow{4}{*}{$32\times 32$}&77.95\tiny{±1.70}&82.25\tiny{±0.19}&60.19\tiny{±7.53}\\
        $\tau$=0.1 (Full-scratch) &&77.41\tiny{±2.21}&82.21\tiny{±0.05}&63.26\tiny{±1.83}\\
        $\tau$=1.0 (Full-scratch) &&78.73\tiny{±1.04}&82.75\tiny{±0.38}&62.60\tiny{±1.08}\\
        $\tau$=2.0 (Full-scratch) &&79.37\tiny{±1.11}&82.28\tiny{±0.27}&62.00\tiny{±0.97}\\
            \hline
        $\tau$=0.01 (Pre-trained) &\multirow{4}{*}{$32\times 32$}&78.06\tiny{±3.26}&82.91\tiny{±0.28}&49.66\tiny{±6.15}\\
        $\tau$=0.1 (Pre-trained) &&80.45\tiny{±0.84}&82.93\tiny{±0.29}&60.17\tiny{±4.45}\\
        $\tau$=1.0 (Pre-trained) &&75.89\tiny{±4.77}&81.46\tiny{±0.67}&61.75\tiny{±0.81}\\
        $\tau$=2.0 (Pre-trained) &&80.35\tiny{±0.44}&75.35\tiny{±2.73}&62.84\tiny{±2.66}\\
    \bhline{1.0pt} 
    \end{tabular*}
    }
\end{table}

Table 2 shows the results of pseudo-label learning for partially-supervised segmentation on the three datasets. 
Original learning in Table 2 indicates the case when the model is trained by a full annotation dataset, and pseudo label learning indicates the case where it is trained by pseudo labels generated by our proposed strategy.
In addition, full-scratch in Table 2 indicates the case where the pre-trained model is not used when we generate pseudo-labels, and pre-trained indicates the case where it is used.
All subsequent training with pseudo labels is done in full scratch.
We fixed 2.0 to the value of $\tau$ because we consider it to be a suitable parameter for all three types of datasets from the results of Table 3.
By using our pseudo label, there was almost no difference in the average DSC compared to results trained using the original overall label images even though the annotation is only in one part of the image.
Particularly, when we used the pre-trained model, the difference of average DSC was within $0.65\%$ for ISBI2012, $0.46\%$ for ssTEM, and $0.40\%$ for iRPE in comparison with a baseline using the original U-Net with the pre-trained model.

Figure 8 shows segmentation results when the model was trained by our pseudo labels in Stage 2 of Figure 5.
As shown in Figure 8, we can confirm that the predicted results are almost identical to the correct label images.
Figure 9 shows the comparison results between the generated pseudo labels by our method in Stage 1 of Figure 5 and the correct labels.
Even though the hard setting, where only part of the annotations is attached in the image, the generated pseudo labels were of largely incorrect quality compared to the original label images.
These results demonstrated that, in the case of cell images, there is no need to attach an annotation to the entire image, and our proposed method can be used to reduce the annotation cost.

\section{Conclusion}

In this study, we proposed a novel one-shot segmentation method and a partially-supervised segmentation method for microscopic cell images. 
Experiments on three different cell image datasets demonstrated that our proposed methods, which is used the pre-trained model and small visual prompt images, can produce highly accurate models even with a small number of training data.
Furthermore, no need to attach an annotation to the entire image by using the proposed strategy and our proposed method can be used to reduce the annotation cost.
However, we were only evaluating simple binary cell image datasets.
Therefore, evaluating whether it is equally effective for multi-class segmentation is our future work.
Additionally, we would like to evaluate zero-shot segmentation using the proposed method.

\section*{Acknowledgements}
This work was supported by JSPS KAKENHI Grant Number 22H04735, and Tateishi Science and Technology Promotion Foundation.

{\small
\bibliographystyle{ieee_fullname}
\bibliography{egbib}
}

\end{document}